\documentclass[final,5p,times,twocolumn]{elsarticle}

\usepackage{amsfonts}

\usepackage{amsmath}
\usepackage{amssymb}
\usepackage{booktabs,multirow}

\usepackage{algpseudocode}
\usepackage[ruled,linesnumbered]{algorithm2e}

\usepackage{bm}
\usepackage{siunitx}
\usepackage{pgf}
\usepackage{tikz}
\usetikzlibrary{bayesnet}
\usetikzlibrary{positioning}
\usetikzlibrary{calc}
\usepackage{wasysym}
\usepackage{makecell}
\usepackage[pdfborder={0 0 0.1}]{hyperref}
\usepackage[capitalise]{cleveref}
\usepackage{caption}
\usepackage{subcaption}
\usepackage{balance}

\definecolor{mpl_propcycle_1}{HTML}{1f77b4}
\definecolor{mpl_propcycle_2}{HTML}{ff7f0e}
\definecolor{mpl_propcycle_3}{HTML}{2ca02c}
\definecolor{mpl_propcycle_4}{HTML}{d62728}

\def\sC{{\mathcal{C}}}

\def\dN{{\mathcal{N}}}

\def\vx{{\bm{x}}}
\def\vu{{\bm{u}}}
\def\vz{{\bm{z}}}
\def\valpha{{\bm{\alpha}}}
\def\vbeta{{\bm{\beta}}}

\def\veps{{\bm{\epsilon}}}

\def\mQ{{\bm{Q}}}

\def\lD{{\mathrm{D}}}

\def\cpx{{p_\mathrm{x}}}
\def\cpy{{p_\mathrm{y}}}
\def\qctx{q_{\mathrm{ctx}}}
\def\qfwd{q_{\mathrm{fwd}}}
\def\aterrfcn{\alpha_{\mathrm{terrain}}}
\def\tterrain{\tau_{\mathrm{terrain}}}
\def\arobot{\valpha_{\mathrm{robot}}}

\SetKwInput{KwInput}{Input}

\journal{Robotics and Autonomous Systems. Copyright (c) the authors. Licensed under \href{https://creativecommons.org/licenses/by-nc-nd/4.0/}{CC BY-NC-ND 4.0}.}

\begin{document}

\begin{frontmatter}



\title{Learning a Terrain- and Robot-Aware Dynamics Model for Autonomous Mobile Robot Navigation}


\author[mpiis]{Jan Achterhold}
\author[mpiis,unifr]{Suresh Guttikonda}
\author[unia]{Jens U. Kreber\corref{cor1}}

\cortext[cor1]{Corresponding author}
\ead{jens.kreber@uni-a.de}
\author[mpiis]{Haolong Li}
\author[mpiis,unia]{Joerg Stueckler}

\affiliation[mpiis]{organization={Embodied Vision Group, Max Planck Institute for Intelligent Systems},
            addressline={Max-Planck-Ring 4}, 
            city={Tuebingen},
            postcode={72070}, 
            state={Baden-Wuerttemberg},
            country={Germany}}

\affiliation[unifr]{organization={University of Freiburg},
            addressline={Georges-Köhler-Allee 51}, 
            city={Freiburg},
            postcode={79110}, 
            state={Baden-Wuerttemberg},
            country={Germany}}

\affiliation[unia]{organization={Intelligent Perception in Technical Systems Group, University of Augsburg},
            addressline={Eichleitnerstrasse 30}, 
            city={Augsburg},
            postcode={86159 }, 
            state={Bavaria},
            country={Germany}}            


\begin{abstract}
Mobile robots should be capable of planning cost-efficient paths for autonomous navigation.
Typically, the terrain and robot properties are subject to variations.
For instance, properties of the terrain such as friction may vary across different locations. 
Also, properties of the robot may change such as payloads or wear and tear, e.g., causing changing actuator gains or joint friction.
Autonomous navigation approaches should thus be able to adapt to such variations.
In this article, we propose a novel approach for learning a probabilistic, \underline{t}errain- and \underline{r}obot-\underline{a}ware forward \underline{\smash{dyn}}amics model (TRADYN) which can adapt to such variations and demonstrate its use for navigation. 
Our learning approach extends recent advances in meta-learning forward dynamics models based on Neural Processes for mobile robot navigation.
We evaluate our method in simulation for 2D navigation of a robot with uni-cycle dynamics with varying properties on terrain with spatially varying friction coefficients.
In our experiments, we demonstrate that TRADYN has lower prediction error over long time horizons than model ablations which do not adapt to robot or terrain variations. 
We also evaluate our model for navigation planning in a model-predictive control framework and under various sources of noise.
We demonstrate that our approach yields improved performance in planning control-efficient paths by taking robot and terrain properties into account.
\end{abstract}








\end{frontmatter}


\section{Introduction}
Autonomous navigation is one of the fundamental functionalities of mobile robots and has been investigated over several decades.
Challenges still remain when robots need to operate in diverse terrain or unstructured environments, for which they require the ability to learn the properties of the environment and to adapt quickly. 
For instance, in weeding in agricultural robotics or search and rescue operations, robots needs to navigate over a wide variety of terrains such as grass, gravel, or mud with varying slope, friction, and other characteristics. 
These properties are often hard to fully and accurately model beforehand~\cite{sonker2021adding}.
Moreover, also properties of the robot itself can be subject to variations due to battery consumption, weight changes, or wear and tear. 
Thus, navigation approaches are needed which allow the robot to adapt to such changes in robot- and terrain-specific properties.

\begin{figure}[thpb]
    \centering
    \includegraphics[width=\linewidth]{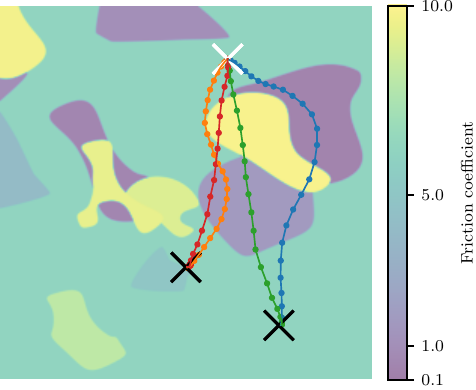}
    \caption{\textbf{Terrain- and robot-aware robot dynamics models for control-efficient navigation}. We propose a novel approach for learning dynamics models for control-cost optimal navigation. Our learned dynamics model can adapt to unobserved  robot properties, such as the mass, and properties of the terrain, such as the friction coefficient which can vary spatially. The above example shows planned paths to two different goals (white/black crosses: start/goal). Our method avoids areas of high friction coefficient and favors areas of low friction coefficient. As the impact of the friction coefficient depends on the mass of the robot, a heavy robot ($m= \SI{4}{\kilo\gram}$, {\color{mpl_propcycle_1}blue}, {\color{mpl_propcycle_2}orange}) takes longer detours to the goal than a light robot ($m= \SI{1}{\kilo\gram}$, {\color{mpl_propcycle_3}green}, {\color{mpl_propcycle_4}red}).}
    \label{fig:tradyn:terrain_aware_motion_planning}
\end{figure}

In this article, we present a novel approach for learning robot- and terrain-aware models for navigation from interaction experience and environment maps.
The key idea of our approach is to learn a deep forward dynamics model which is conditioned on learned context variables that model robot- and terrain-specific properties. 
The context variable modeling robot-specific properties is inferred online from observed state transitions.
The terrain context features are extracted from an a-priori known environment map and additionally used to condition the dynamics model.

We evaluate our approach in a 2D dynamics simulation. 
The mobile robot is modeled as a point mass moving with unicycle driving dynamics.
The dynamics depend on a couple of robot- and terrain-specific parameters.
Terrains are defined by regions between which the friction properties vary, while robot properties such as mass can also change between environments.
In our experiments, we demonstrate that our learned context-aware dynamics model can capture the varying robot and terrain properties well and supercedes a dynamics model without context-awareness in prediction accuracy and planning performance.
We also provide insights into the effect of various sources of noise in actions and observations on the performance of our learning approach.

To summarize we make the following contributions:
\begin{enumerate}
    \item We propose a probabilistic deep forward dynamics model which can adapt to robot- and terrain-specific properties that influence the mobile robot's dynamics. 
    \item We demonstrate in a 2D simulation environment that these adaptation capabilities are crucial for the predictive performance of the dynamics model.
    We assess the influence of noise on the performance of our model for prediction and motion planning.
    \item The learned context-aware dynamics model is used for robot navigation by model-predictive control. This way, efficient paths can be planned that take robot and terrain properties into account (see \cref{fig:tradyn:terrain_aware_motion_planning}).
\end{enumerate}

This article extends our work in~\cite{guttikonda2023_etcnav} by including an assessment of the impact of various noise sources such as action and sensor noise on prediction and planning performance of our approach.
   
\section{Related work}

Autonomous mobile robot navigation in diverse terrain and robot conditions has been subject to research over the last decades.
Recent approaches are typically learning-based to infer terrain properties from sensory and map information. 

\subsection{Terrain-aware navigation}

Early approaches have used pretrained semantic segmentation to determine terrain categories~\cite{valada2017adapnet,yang1028unifying}.
The information can be used to only navigate on segments of traversable terrain.
Path planners typically rely on some form of costs for traversing specific terrain types. 
In Zhu et al.~\cite{zhu2019offroad} the terrain traversal costs are learned from human demonstrations by inverse reinforcement learning. 
TerrainNet~\cite{meng2023_terrainnet} learns a cost function for traversing diverse and uneven surfaces.
These methods, however, do not learn the dynamical properties of the robot on the terrain classes explicitly like our method.

A predictive model of future events for navigation planning such as collisions, bumpiness and position is learned in BADGR~\cite{kahn2021badgr}.
The prediction is inferred from the current RGB image and control actions and the model is trained supervised from sample trajectories in the same environment.
A dynamics model of driving behavior for model-predictive control is learned in Grigorescu et al.~\cite{grigorescu2021lvdnmpc}.
The model is conditioned on input camera images for state estimation and predictions. 
Different to our approach, however, the model does not explicitly capture a variety of terrain- and robot-specific properties jointly.
The approach in Siva et al.~\cite{siva2021enhancing} learns offsets from the actual robot behavior based on multimodal terrain features which are extracted from camera, LiDAR, and IMU data. 
For high-speed motion planning on unstructured terrain,, the approach in Xiao et al.~\cite{xiao2021learning} learns an inverse kinodynamics model from inertial measurements.
Sikand et al.~\cite{sikand2022visual} propose an approach for preference-aware path planning that embeds visual features of terrain which have similar traversability properties close in feature space using contrastive learning.
Different to our approach, the above approaches do not distinguish terrain- and robot-specific properties explicitly in a dynamics model nor model them concurrently.

Closely related to our approach, Vertens et al.~\cite{vertens2023_deepdynlatent} propose to learn latent embeddings that model surface properties.
These latents are stored in a 2D map of the environment and parameterize a learned neural dynamics model that predicts the next state based on the last state and the control input.
However, this approach does not model robot-specific parameters explicitly.

\subsection{Learning of dynamics models for navigation}

Several approaches for learning action-conditional dynamics models have been proposed in the machine learning and robotics literature in recent years.
In the seminal work PILCO~\cite{deisenroth2011pilco}, Gaussian processes are used to learn to predict subsequent states, conditioned on actions.
The approach is demonstrated for balancing and swinging up a cart-pole.
Several approaches learn latent embeddings of images and predict future latent states conditioned on actions using recurrent neural networks~\cite{lenz2015deepmpc,oh2015action,finn2016unsupervised,hafner2019learning}.
The models are used in several of these works for model-predictive control and planning.
Learning-based dynamics models are also popular in model-based reinforcement learning (see, e.g.,~\cite{nagabandi2018neural}).
Shaj et al.~\cite{shaj2020acrnn} propose action-conditional recurrent Kalman networks which implement observation and action-conditional state-transition models in a Kalman filter with neural networks.
In Gibson et al ~\cite{gibson2023_multistep}, a recurrent neural network architecture is trained to predict the movement of the vehicle based on previous states for offroad driving.
Related are also approaches which calibrate parameters of motion models online with state estimation~\cite{weydert2012_ekfvoparameterid} or SLAM~\cite{li2022_kinvio}.

\subsection{Meta-learning families of dynamics models}

The above recurrent neural network approaches can aggregate context from previous experience in the latent state of the network.
Other approaches can incorporate an arbitrarily sized context observation set to infer a context variable. 
Lee et al.~\cite{lee2020context} propose to parametrize dynamics models in model-based reinforcement learning on a context embedding which is inferred from past experience.
Achterhold and Stueckler~\cite{achterhold2021explore} learn context-conditional dynamics models using Neural Processes~\cite{garnelo2018neural} to infer a probabilistic belief of the context variable.
Rapid motor adaptation~\cite{kumar2021_rma} (RMA) learns a controller for legged locomotion in simulation which is parametrized with a learned neural embedding on a large number of simulation parameter variations. 
At the same time, an inference network is learned to estimate the embedding vector from robot motion data which is reused later for the real robot for sim-to-real transfer.
Our approach is based on the context-conditional dynamics model learning approach in~\cite{achterhold2021explore}.
It uses Neural Processes~\cite{garnelo2018neural} to infer robot-specific parameters as the distribution of a context variable in a learned context embedding.

\section{Background}
\label{sec:background}

Our approach extends the context-conditional probabilistic neural dynamics model of \emph{Explore the Context} (EtC~\cite{achterhold2021explore}) for robot navigation.
EtC assumes that the dynamical system to be modelled can be formulated by a discrete-time state-space model
\begin{equation}
    \label{eq:tradyn:etc_assumption}
    \vx_{n+1} = f( \vx_{n}, \vu_{n}, \valpha ) + \veps_n,~\veps_n \sim \dN( 0, \mQ_n ),
\end{equation}
with Markov properties for the state transition.
In this model,~$\vx_{n}$ is the state at time step~$n$,~$\vu_{n}$ is the control input, and~$\valpha$ is a latent, i.e., unobserved, variable which parametrizes the dynamics, e.g., comprising robot or terrain parameters. 
The state transitions are assumed to be stochastic with Gaussian additive noise~$\veps_n$ of zero mean and diagonal covariance~$\bm{Q}_n$.
Besides $\valpha$ and its representation, also the dynamics~$f$ is unknown and needs to be learned from data. 

EtC learns an approximation of the dynamics using a recurrent neural forward dynamics model~$\qfwd$. 
The environment-specific properties~$\valpha$ are incorporated into the learned model by conditioning it on a context variable~$\vbeta \in \mathbb{R}^B$ whose representation is learned concurrently.
A probability density on~$\bm{\beta}$ is inferred from a \emph{context} set 
\begin{equation}
\sC^\alpha = {\{(\vx^{(k)}, \vu^{(k)}, \vx^{(k)}_+)\}}_{k=1}^K.
\end{equation}
of~$K$ observed state transitions ($\vx_+ \leftarrow \vx,\vu$) which are assumed to follow the modeled dynamics and are obtained by interaction experience in the environment.

A \emph{context encoder}~$\qctx(\vbeta \mid \sC^\alpha)$ is trained to infer the probability density on~$\vbeta$ from the context set. 
Training is performed to predict target rollouts $\lD^\alpha = [\vx_0, \vu_0, \vx_1, \vu_1, \ldots, \vu_{N-1}, \vx_{N}]$. 
Context set and target rollout are from the same environment instance parametrized by~$\bm{\alpha}$.
The learning objective maximizes the marginal log-likelihood of the target rollout given the context set,
\begin{equation}
   \label{eq:tradyn:mll}
   \log p( \lD^\alpha \mid \sC^\alpha ) = \log \int p( \lD^\alpha \mid \vbeta ) \, p( \vbeta \mid \sC^\alpha ) \, d\vbeta,
\end{equation}
where marginalization is performed over the latent context variable~$\vbeta$.
To learn a model over differently parametrized environments, we aim to maximize $\log p( \lD^\alpha \mid \sC^\alpha )$ in expectation over the distribution of environments~$\Omega_\alpha$, and a sample set of target rollouts and corresponding context sets $\Omega_{\lD^\alpha,\sC^\alpha}$, i.e.,
\begin{equation}
    \mathbb{E}_{\valpha \sim \Omega_\alpha, (\lD^\alpha,\sC^\alpha) \sim \Omega_{\lD^\alpha,\sC^\alpha}} \left[ \log p( \lD^\alpha \mid \sC^\alpha ) \right].
\end{equation}
The term $p( \lD^\alpha \mid \vbeta )$ is modeled by single-step and multi-step prediction factors and reconstruction factors derived from the learned forward dynamics model $\qfwd \left( \vx^{}_{n} \mid \vx^{}_{0}, \vu^{}_{0:n-1}, \vbeta \right)$.
The distribution $p(\vbeta \mid \sC^\alpha)$ is learned by $\qctx(\vbeta \mid \sC^\alpha)$.

The forward dynamics model is implemented as a recurrent neural network using gated recurrent units (GRU,~\cite{cho2014learning}).
The initial state~$\bm{x}_0$ is encoded into the initial hidden state~$\bm{z}_0$ of the GRU. 
The GRU gets control input~$\vu_n$ and context variable~$\vbeta$ as input after encoding them into feature vectors
\begin{align}
\bm{z}_0 &= e_{x}(\vx_{0}) \\
\mathbf{z}_{n+1} &= \operatorname{GRU} \left( \bm{z}_n, [e_{u}(\vu_n), e_{\beta}(\vbeta) ] \right) \label{eq:tradyn:gru}
\end{align}
where~$e_{x}$,~$e_{u}$, and~$e_{\beta}$ are encoders implemented as multi-layer perceptrons (MLPs). 
The GRU predicts latent states~$\mathbf{z}_{n}$ which are decoded into a Gaussian state distribution in the original state space
\begin{equation}
    \begin{split}
        \vx_{n} &\sim \mathcal{N} \left( d_{x,\mu}(\mathbf{z}_{n}), d_{x,\sigma^2}(\mathbf{z}_{n}) \right)
    \end{split}
\end{equation}
using MLPs~$d_{x,\mu}$, $d_{x,\sigma^2}$.

Input to the the context encoder is the context set~$\sC^\alpha$ which is a set of state-action-state transitions with variable size $K$.
The context encoder first encodes each transition in the context set independently using a transition encoder~$e_{\mathrm{trans}}$.
The encoded transitions are then aggregated using a dimension-wise permutation-invariant max operation into the latent variable~$\bm{z}_{\beta}$.
A Gaussian distributed belief over the context variable~$\vbeta$ is predicted from the aggregated encodings
\begin{equation}
    \qctx( \vbeta \mid \sC^\alpha ) = \mathcal{N}\left( \vbeta; d_{\beta,\mu}( \vz_\beta ), \operatorname{diag}( d_{\beta,\sigma^2}( \vz_\beta ) ) \right)
\end{equation} 
with MLP decoders~$d_{\beta,\mu}$,~$d_{\beta,\sigma^2}$.
To guarantee that the predicted variance is positive and decreases monotonically with increasing the number of context observations, the network~$d_{\beta,\sigma^2}$ squashes the negated output of an MLP which has non-negative weights and activations and is applied to $\vz_\beta$ through a Softplus activation function.

Similar to Neural Processes~\cite{garnelo2018neural}, to obtain a tractable loss, the marginal log likelihood in \cref{eq:tradyn:mll} is approximately bounded using the evidence lower bound (ELBO, c.f.~\cite{le2018empirical})
\begin{multline}
    \label{eq:tradyn:elbo}
    \log p( \lD^\alpha \mid \sC^\alpha ) \gtrapprox \mathbb{E}_{\vbeta \sim  \qctx( \vbeta \mid \lD^\alpha \cup \sC^\alpha )} \left[ \log p( \lD^\alpha \mid \vbeta ) \right]\\
    - \lambda_{\mathit{KL}} \operatorname{KL}\left( \qctx( \vbeta \mid  \lD^\alpha \cup \sC^\alpha )  \, \| \, \qctx( \vbeta \mid \sC^\alpha ) \right).
\end{multline}
Dynamics model and context encoder are trained jointly by maximizing the ELBO by stochastic gradient ascent on empirical samples of corresponding target rollouts and context sets, obtained on a training set of environments. 

At test time, the dynamics model~$\qfwd(\vx^{}_{n} \mid \vx^{}_{0}, \vu^{}_{0:N-1}, \vbeta)$ can be adapted to a particular environment instance $\valpha$ (called \emph{calibration}) by inferring~$\vbeta$ from collected context observations using~$\qctx(\vbeta \mid \sC^\alpha)$.

  \begin{figure}
        \includegraphics[scale=1]{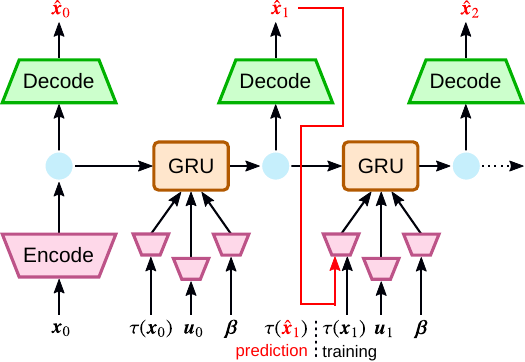}
      \caption{Terrain- and robot-aware forward dynamics model (TRADYN).
      We embed the initial robot state~$\bm{x}_0$ as hidden state of a gated recurrent unit (GRU). The GRU predicts the next state in the latent space for which it receives context encoding~$\bm{\beta}$ and embeddings of action~$\bm{u}$ and terrain observation~$\bm{\tau}$ as additional inputs. Latent states are decoded into Gaussian distributions on the robot's next state. While during training the actual terrain observation~${\tau}(\bm{x}_n)$ is used, during prediction, the map~${\tau}$ is queried at predicted robot locations~${\tau}({\color{red}\bm{\hat{x}}_n})$. See~\cref{sec:tradyn:method} for details.
      }
      \label{fig:tradyn:model}
  \end{figure}

\section{Method}
\label{sec:tradyn:method}
EtC assumes that the dynamics varies across different instances of the environment and can be captured in a latent variable~$\valpha$ (see \cref{eq:tradyn:etc_assumption}) per environment.
In our robot navigation problem, the terrain varies across locations in each environment, while the robot-specific parameters are again assumed to be a global latent variable~$\arobot$ in the respective environment.
While in principle, one could try to model the terrain layout by a latent variable~$\aterrfcn$, we model the terrain latent location-dependent by a state-dependent function $\aterrfcn(\vx_n)$.

\subsection{Terrain- and Robot-Aware Dynamics Model}
In our approach, we assume that the robot motion in an environment can be described by the following dynamical system
\begin{equation} \label{eq:dynamics_assumption}
    \vx_{n+1} = f( \vx_n, \vu_n, \arobot, \aterrfcn(\vx_n)) + \veps_n
\end{equation}
with~$\veps_n \sim \mathcal{N}( 0, \mQ_n)$ as in \cref{eq:tradyn:etc_assumption}.
Here,~$\vx_n$ refers to the robot state at time step~$n$,~$\vu_n$ are the control inputs,~$\arobot$ parametrizes latent properties of the robot (mass, actuator gains), and~$\aterrfcn(\vx_n)$ parametrizes the spatially dependent terrain properties (e.g., friction). 
We also assume that the terrain parameters~$\aterrfcn$ cannot be directly observed.
Instead, we assume that a map of \emph{terrain features}~$\tterrain(\vx_n)$ is known, which can be queried at any $\bm{x}_n$, and that~$\aterrfcn(\vx_n)$ can be inferred from~$\tterrain(\vx_n)$.
For example,~$\tterrain$ may map visual terrain observations that relate to friction coefficients. 

Consequently, we condition the learned forward dynamics model on observed terrain features~$\bm{\tau}_{0:n-1}$ for terrain-awareness and on the latent variable~$\bm{\beta}$ which captures robot-specific variations due to~$\arobot$, i.e.,
\begin{equation}
    \bm{\hat{x}}_{n} \sim \qfwd(\bm{x}_{n} \mid \bm{x}_{0}, \bm{u}_{0:n-1}, \bm{\beta}, \bm{\tau}_{0:n-1}).
\end{equation}
The latent variable~$\bm{\beta}$ still needs to be inferred from context observations, i.e., observed state transitions in the environment.
The terrain features~$\bm{\tau}_{0:n-1}$ need to be obtained differently for training and prediction. 
During training, we retrieve~${\tau}$ at ground-truth states, i.e.~$\bm{\tau}_{0} = {\tau}(\bm{x}_{0}), \bm{\tau}_{1} = {\tau}(\bm{x}_{1})$, etc. 
For prediction, we do not have access to ground-truth states, and obtain $\bm{\tau}_{1:n-1}$ at positions predicted by the learned forward dynamics model, i.e., $\bm{\tau}_{1} = {\tau}(\bm{\hat{x}}_{1}), \bm{\tau}_{2} = {\tau}(\bm{\hat{x}}_{2})$, etc. 

We extend the network architecture of EtC as follows to incorporate terrain-specific properties: 
An MLP~$e_\tau$ encodes the terrain feature $\bm{\tau}$ at the position of the corresponding time step and the encoding is passed as additional input to the GRU, so that~\cref{eq:tradyn:gru} becomes
\begin{equation}
    \mathbf{z}_{n+1} = \operatorname{GRU} \left( \bm{z}_0, [e_{\tau}(\boldsymbol{\tau}_n), e_{u}(\vu_n), e_{\beta}(\vbeta) ] \right).
\end{equation}
The terrain features are also included to accompany the respective states in the context set, i.e.,
\begin{equation}
\begin{aligned}
\sC^\alpha = {\{(\vx^{(k)}, {\tau}(\vx^{(k)}),  \vu^{(k)}, \vx_{+}^{(k)}, {\tau}(\vx_{+}^{(k)}))\}}_{k=1}^K.
\end{aligned}
\end{equation}
\cref{fig:tradyn:model} illustrates our model.
Network architecture details are provided in the section~\ref{sec:network_architecture}.

\subsection{Path Planning and Motion Control}
\label{sec:tradyn:pathplanning}

We use TRADYN in a model-predictive control setup. The model $\qfwd$ yields state predictions $\bm{\hat{x}}_{1:H}$ for an initial state $\bm{x_0}$ and controls $\bm{u}_{0:H-1}$, for planning horizon $H$. For calibration, i.e., inferring $\bm{\beta}$ from a context set $\sC$ with the context encoder $\qctx$, calibration transitions are collected on the target environment prior to planning. This allows adapting to varying robot parameters. The predictive terrain feature lookup (see \cref{fig:tradyn:model}) with ${\tau}(\vx)$ allows adapting to varying terrains. 
We use the Cross-Entropy Method (CEM~\cite{rubinstein1999cross}) for planning.
We aim to reach the target position with minimal throttle control energy, given by the sum of squared throttle commands during navigation. This gives rise to the following planning objective, which penalizes high throttle control energy and a deviation of the robot's terminal position to the target position $\bm{p}^*$:
\begin{multline}
    \label{eq:tradyn:cost}
     J(\bm{u}_{0:H-1}, \bm{\hat{x}}_{1:H}) = \\ \frac{1}{2} \sum_{n=0}^{H-1} u_{\mathrm{throttle}, n}^2 + \lambda_{\mathrm{dist}} \left( ||  \bm{\hat{p}}_{H} - \bm{p}^* ||_2^2 - \gamma_{\mathrm{dist}} \right).
\end{multline}

We observe that large terrain and thus friction coefficient variations renders a global setting of $\lambda_{\mathrm{dist}}$ for weighting the control cost term problematic.
Depending on the terrain, different trade offs between the cost terms are needed.
In our CEM implementation, we thus normalize the distance term in \cref{eq:tradyn:cost} using $\lambda_{\mathrm{dist}}, \gamma_{\mathrm{dist}}$ to zero mean and unit variance over all CEM candidates.
At each step, model-predictive control only applies the first action and replans from the next, resulting state using a receding horizon scheme.

\section{Experiments}

\begin{figure}[tb]
    \centering
    \includegraphics[scale=1]{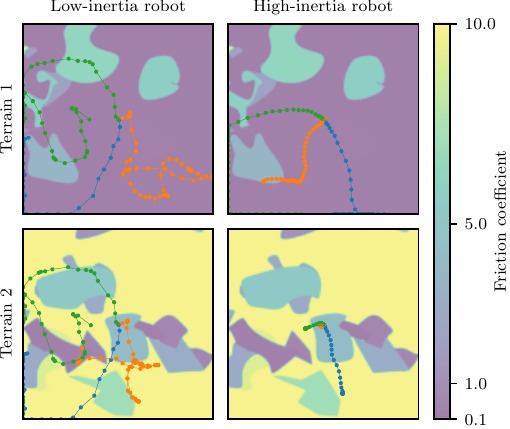}
    \caption{Exemplary rollouts (length 50) on two different terrain layouts (rows) and for two exemplary robot configurations (low-inertia, high-inertia) (columns). Rollouts start from the center; actions are sampled time-correlated. The low-inertia robot has minimal mass $m=1$ and maximal control gains $k_\mathrm{throttle}=1000$, $k_\mathrm{steer}=\pi/4$. The high-inertia robot has maximal mass $m=4$ and minimal control gains $k_\mathrm{throttle}=500$, $k_\mathrm{steer}=\pi/8$. Equally colored trajectories ({\scriptsize \color{mpl_propcycle_1} \newmoon}, {\scriptsize \color{mpl_propcycle_2} \newmoon}, {\scriptsize \color{mpl_propcycle_3} \newmoon}) correspond to identical sequences of applied actions. See \cref{sec:tradyn:simenv} for details.}
    \label{fig:tradyn:rollouts}
\end{figure}

\begin{figure}[tb]
    \centering
    \includegraphics[scale=1]{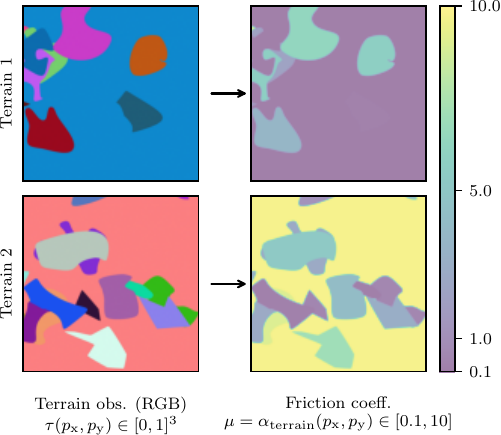}
    \caption{Relationship of RGB terrain features $\bm{\tau}$ (left column) to friction coefficient $\mu$ (right column). See \cref{sec:tradyn:terrainlayouts} for details.}
    \label{fig:tradyn:terrains}
\end{figure}

\begin{figure*}[tb!]
    \centering
    \includegraphics[scale=1]{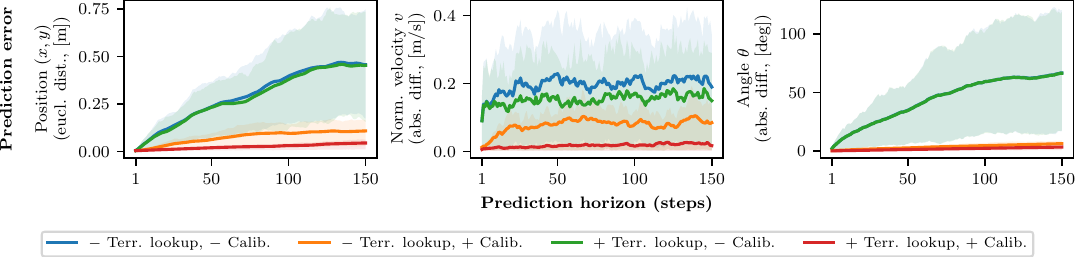}
    \caption{Prediction error evaluation for the proposed model and its ablations (no terrain lookup / no calibration), plotted over the prediction horizon (number of prediction steps). From left to right: Positional error (euclidean distance), velocity error (absolute difference), angular error (absolute difference). Depicted are the mean and 20\%, 80\% percentiles over 150 evaluation rollouts for 5 independently trained models per model variant. Our approach with terrain lookup and calibration clearly outperforms the other variants in position and velocity prediction (left and center panel). For predicting the angle (right panel), terrain friction is not relevant, which is why the terrain lookup brings close to no advantage. However, calibration is important for accurate angle prediction. See \cref{sec:tradyn:eval_prediction} for details.}
    \label{fig:tradyn:unicycle_prediction}
\end{figure*}

\begin{figure}[tb!]
    \centering
    \includegraphics[scale=1]{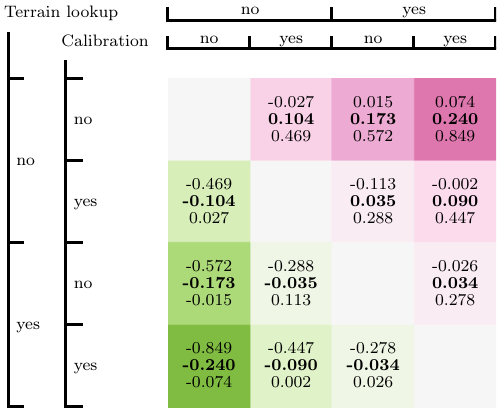}\\
    \caption{Comparison of model variants with and without terrain lookup and calibration. $E^v_{k,i}$ denotes the throttle control energy for method $v$ on navigation task $k \in \{1, \ldots, 150\}$ for a trained model with seed $i \in \{1, \ldots, 5\}$. We show statistics (20\% percentile, median, 80\% percentile) on the set of pairwise comparisons of control energies $\{ 
    E^{\mathrm{row}}_{k,i_1} - E^{\mathrm{col}}_{k,i_2} \mid \forall k \in \{1,\ldots,K\}, i_1 \in \{1,\ldots,5\}, i_2 \in \{1,\ldots,5\}
    \}$. Significant ($p<0.05$) results are printed \textbf{bold} (see \cref{sec:tradyn:eval_planning}). Exemplarily, both performing terrain lookup and calibration (last row) yields navigation solutions with significantly lower throttle control energy (negative numbers) compared to all other methods (columns). See \cref{sec:tradyn:eval_planning} for details.
    }
    \label{fig:tradyn:unicycle_planning_gain}
\end{figure}

\begin{figure}[tb!]
    \centering
    \includegraphics[scale=1]{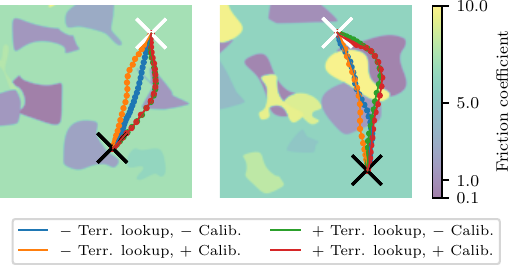}\\
    \vspace{1em}
    {\footnotesize
    \begin{tabular}{lcccc} 
    \toprule
    & \multicolumn{2}{c}{left terrain} & \multicolumn{2}{c}{right terrain} \\
     \cmidrule(lr){2-3} \cmidrule(lr){4-5}
    Variant & \makecell{Thr.\ ctrl.\\energy} & \makecell{Target\\dist [\SI{}{mm}]} & \makecell{Thr.\ ctrl.\\energy} & \makecell{Target\\dist [\SI{}{mm}]}\\
    \midrule 
        {\scriptsize \color{mpl_propcycle_1} \newmoon} $-$T, $-$C & 3.40 & 10.34& 3.61 & 6.92 \\
        {\scriptsize \color{mpl_propcycle_2} \newmoon} $-$T, $+$C & 2.65 & 8.39 & 2.91 & 5.97 \\
        {\scriptsize \color{mpl_propcycle_3} \newmoon} $+$T, $-$C & 2.39 & 5.62 & 1.92 & 5.28 \\
        {\scriptsize \color{mpl_propcycle_4} \newmoon} $+$T, $+$C & 2.11 & 4.20 & 1.64 & 3.98 \\
    \bottomrule
    \end{tabular}
    }
    \caption{
    Exemplary navigation trajectories and their associated throttle control energy and final distance to the target (see table). The robot starts at the white cross, the goal is marked by a black cross. With terrain lookup ({\scriptsize \color{mpl_propcycle_3} \newmoon} +T, -C and  {\scriptsize \color{mpl_propcycle_4} \newmoon} +T, +C), our method circumvents areas of high friction coefficient (i.e., high energy dissipation), resulting in lower throttle control energy (see table). Enabling calibration (+C) further reduces throttle control energy. See \cref{sec:tradyn:eval_planning} for details.}
    \label{fig:tradyn:unicycle_planning_examples}
\end{figure}

\begin{figure}[tb!]
    \centering
    \includegraphics[scale=1]{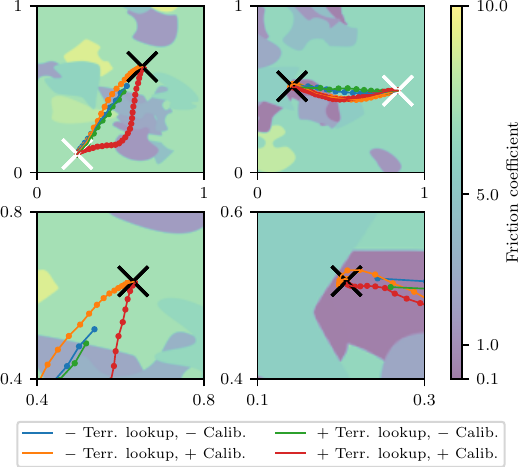}\\
    \caption{Failure cases for the non-calibrated models. The top row shows the full terrain of extent $[0, 1]$\,\SI{}{\meter}. The bottom row is zoomed around the goal. In the cases shown, planning with the non-calibrated models does not succeed in reaching the goal marked by the black cross within the given step limit of $50$ steps, in contrast to the calibrated models. See \cref{sec:tradyn:eval_planning} for details.}
    \label{fig:tradyn:unicycle_planning_fail}
\end{figure}

We structure the experiments section as follows:
In~\cref{sec:tradyn:simenv}, we describe the properties of the simulation environment in which we conduct our experiments.
We evaluate our method on a simulated unicycle robot with varying parameters in a bounded 2D environment with varying terrain properties.
In sections~\ref{sec:model_training} to~\ref{sec:model_ablation}, we detail the architecture of our employed model and the training process.
To demonstrate the efficacy of our proposed method, we evaluate it together with several ablations:
We train a model variant that has no access to a terrain map for lookup, but is only fed the current terrain feature as observation.
Also, we evaluate both variants when provided with an informative context set, which we then call \textit{calibrated} and also with an empty context set.
This way, we can closely study the effect of context on the model.

Concretely, we seek to answer the following research questions:
\begin{itemize}
    \item How does providing global robot-specific and local terrain-specific environment context to a learned dynamics model impact its trajectory prediction performance?
    How does it impact the model's usefulness in planning motion to a goal location?\\
    To this end, we evaluate our proposed model and its ablations regarding robot state prediction error (\cref{sec:tradyn:eval_prediction}) and in a CEM planning scenario (\cref{sec:tradyn:eval_planning}).
    We conduct these experiments on deterministic environments and with no forms of noise present.
    \item How do different forms of noise, specifically stochastic environment dynamics and noisy observations, impact the planning and prediction performance of our proposed approach?
    We conduct similar experiments regarding prediction and planning for all model variants with various forms of noise present in the environment in \cref{sec:noise_experiments}.
\end{itemize}

\subsection{Simulation environment}
\label{sec:tradyn:simenv}

\subsubsection{Simulated Robot Dynamics} \label{sec:sim_robot_dynamics}

We perform experiments in a 2D simulation with a unicycle-like robot setup where the continuous time-variant 2D dynamics with position $\mathbf{p} = {[\cpx, \cpy]}^\top$, orientation $\varphi'$, and directional velocity $v$, for control input $\bm{u} = {[u_\mathrm{throttle}, u_\mathrm{steer}]}^\top \in {[-1, 1]}^2$, are given by

\begin{equation}
    \label{eq:tradyn:continuous_time_turtlebot_dynamics}
    \begin{aligned}
        \dot{\bm{p}}(t) & = {\begin{bmatrix} \cos\varphi' & \sin\varphi' \end{bmatrix}}^{\top} {v}(t), \\
        \dot{{v}}(t) & = \frac{1}{m} (F_\mathrm{throttle} + F_\mathrm{fric}), \\
        F_\mathrm{throttle} & = (u_\mathrm{throttle} + \epsilon_\mathrm{throttle}) \: k_\mathrm{throttle}, \\
        F_\mathrm{fric} & = -\operatorname{sign}(v(t)) \: \mu \: m \: g, \\
        \epsilon_\mathrm{throttle} &\sim \dN(0, \sigma_a^2). \\
    \end{aligned}
\end{equation}

As our method does not use continuous-time observations, but only discrete-time samplings with stepsize $\Delta_T = \SI{0.01}{\second}$, we approximate the state evolution between two timesteps as follows.
First, we apply the change in angle as 
$\varphi' = \varphi(t + \Delta_T) = \varphi(t) + (u_\mathrm{steer} + \epsilon_\mathrm{steer}) \: k_\mathrm{steer}$
where $\epsilon_\mathrm{steer} \sim \dN(0, \sigma_a^2)$.
We then query the terrain friction coefficient $\mu$  at the position $\bm{p}(t)$.
With the friction coefficient $\mu$ and angle $\varphi'$, we compute the evolution of position and velocity with \cref{eq:tradyn:continuous_time_turtlebot_dynamics}.
The existence of the friction term in \cref{eq:tradyn:continuous_time_turtlebot_dynamics} requires an accurate integration, which is why we solve the initial value problem in \cref{eq:tradyn:continuous_time_turtlebot_dynamics} numerically using an explicit Runge Kutta (RK45) method, yielding ${\bm{p}}(t+\Delta_T)$ and ${{v}}(t+\Delta_T)$.
To avoid discontinuities, we represent observations of the above system as $\bm{x}(t) = {[\cpx(t), \cpy(t), v(t), \cos \varphi(t), \sin \varphi(t)]}^\top$. We use $g=\SI{9.81}{\metre\per\square\second}$ as gravitational acceleration. Positions $\bm{p}(t)$ are clipped to the range $[0, 1] \: \SI{}{\metre}$; the directional velocity $v(t)$ is clipped to $[-5, 5] \: \SI{}{\metre\per\second}$. The friction coefficient $\mu = \aterrfcn(\cpx, \cpy) \in [0.1, 10]$ depends on the terrain layout $\aterrfcn$ and the robot's position. 
The mass $m$ and control gains $k_\mathrm{throttle}, k_\mathrm{steer}$ are robot-specific properties, we refer to \cref{table:tradyn:robot_specific_properties_setup} for their value ranges.

We consider stochastic dynamics induced by the noisy execution of actions, denoted by $\epsilon_\mathrm{throttle}$ and $\epsilon_\mathrm{steer}$.
In each timestep, both are sampled from a zero-mean Gaussian with identical standard deviation $\sigma_a$.
Note that this form of noise needs to be approximated with the noise model of EtC (\cref{eq:tradyn:etc_assumption} and \cref{eq:dynamics_assumption}) by time-dependent noise.
For the following experiments, if not stated otherwise, we consider deterministic dynamics with $\sigma_a = 0$ and thus $\epsilon_\text{throttle} = \epsilon_\text{steer} = 0$.
Stochastic dynamics are evaluated in~\cref{sec:noise_experiments}.
In addition, we evaluate the effect of noisy state observations in~\cref{sec:noise_experiments}.

\subsubsection{Terrain layouts}
\label{sec:tradyn:terrainlayouts}
To simulate the influence of varying terrain properties on the robots' dynamics, we programmatically generate 50 terrain layouts for training the dynamics model and 50 terrain layouts for testing (i.e., in prediction- and planning evaluation). For generating terrain $k$, we first generate an unnormalized feature map $\hat{\tau}^{(k)}$, from which we compute $\aterrfcn^{(k)}$ and the normalized feature map ${\tau}^{(k)}$.
The unnormalized feature map is represented by a 2D RGB image of size $460\,\mathrm{px} \times 460\,\mathrm{px}$.
For its generation, first, a background color is randomly sampled, followed by sequentially placing randomly sampled patches with cubic bezier contours.
The color value $(r, g, b) \in {\{0, \ldots, 255\}}^3$ at each pixel maps to the friction coefficient $\mu = \alpha_\mathrm{terrain}(\cpx,\cpy)$ through bitwise left-shifts $\ll$ as
\begin{equation}
    \label{eq:tradyn:tau_to_mu}
    \begin{split}
    & \eta =  {((r \ll 16) + (g \ll 8) + b)}/{(2^{24}-1)} \\
    & \alpha_\mathrm{terrain}(\cpx,\cpy) = 0.1 + (10-0.1)\eta^2.
    \end{split}
\end{equation}
The agent can observe the normalized terrain color $\tau^{(k)}(\cpx, \cpy) \in {[0, 1]}^3$ with $\tau^{(k)}(\cpx, \cpy) = \hat{\tau}^{(k)}(\cpx, \cpy) / 255$, and can query $\tau^{(k)}(\cpx, \cpy)$ at arbitrary $\cpx, \cpy$.
The simulator has direct access to $\mu = \aterrfcn^{(k)}(\cpx, \cpy)$.
We denote the training set of terrains $\mathcal{A}_\mathrm{train} = \{ \aterrfcn^{(k)} \mid k \in \{1,\ldots,50\} \}$ and the test set of terrains $\mathcal{A}_\mathrm{test} = \{ \aterrfcn^{(k)} \mid k \in \{51,\ldots,100\} \}$.
We refer to \cref{fig:tradyn:terrains} for a visualization of two terrains and the related friction coefficients.

The robot's dynamics depends on the terrain $\aterrfcn$ as it is the position-dependent friction coefficient, and the robot-specific parameters $\arobot = (m, k_\mathrm{throttle}, k_\mathrm{steer})$. A fixed tuple $(\aterrfcn,\arobot)$ forms an environment \emph{instance}.

\subsubsection{Trajectory generation}
We require the generation of trajectories at multiple places of our algorithm for training and evaluation: To generate training data, to sample candidate trajectories for the cross-entropy planning method, to generate calibration trajectories, and to generate trajectories for evaluating the prediction performance. One option would be to generate trajectories by independently sampling actions from a Gaussian distribution at each time step. However, this Brownian random walk significantly limits the space traversed by such trajectories~\cite{pinneri2020sample}. To increase the traversed space,~\cite{pinneri2020sample} propose to use time-correlated (colored) noise with a power spectral density $\operatorname{PSD}(f) \propto \frac{1}{f^\omega}$, where $f$ is the frequency. We use $\omega=0.5$ in all our experiments.

\subsubsection{Exemplary rollouts} 
\label{par:tradyn:exemplaryrollouts}
We visualize exemplary rollouts on different terrains and with different robot parametrizations in \cref{fig:tradyn:rollouts}. We observe that both the terrain-dependent friction coefficient $\mu$, as well as the robot properties, have a significant influence on the shape of the trajectories, highlighting the importance of a model to be able to adapt to these properties.

\begin{table}[t]
\begin{center}
\caption{Robot-specific properties.}
\label{table:tradyn:robot_specific_properties_setup}
\begin{tabular}{lcc}
\toprule
\multicolumn{1}{l}{ Property} &\multicolumn{1}{c}{ Min.} &\multicolumn{1}{c}{Max.} \\
\midrule
Mass $m$ [kg]          & 1    & 4\\
Throttle gain $k_\mathrm{throttle}$          &500  &1000\\
Steer gain $k_\mathrm{steer}$      & $\pi$/8   & $\pi$/4\\
\bottomrule
\end{tabular}
\end{center}
\vspace{-1em}
\end{table}

\subsection{Model training} \label{sec:model_training}

We train our proposed model on a set of precollected trajectories on different terrain layouts and robot parametrizations. 
First, we sample a set of $10000$ unique terrain layout / robot parameter settings to generate training trajectories. For validation, a set of additional 5000 settings is used. On each setting, we generate two trajectories, used later during training to form the target rollout $\lD^\alpha$ and context set $\sC^\alpha$, respectively.
Terrain layouts are sampled uniformly from the training set of terrains, i.e., $\mathcal{A}_\mathrm{train}$. Robot parameters are sampled uniformly from the parameter ranges given in \cref{table:tradyn:robot_specific_properties_setup}.
The robots' initial state $\bm{x_0} = {[p_{\mathrm{x},0}, p_{\mathrm{y},0}, v_0, \varphi_0]}^\top$ is uniformly sampled from the ranges $p_{\mathrm{x},0}, p_{\mathrm{y},0} \in [0, 1)$, $v_0 \in [-5, 5)$, $\varphi_0 \in [0, 2\pi)$. Each trajectory consists of $100$ applied actions and the resulting states. We use time-correlated (colored) noise to sample actions (see previous paragraph). 
We follow the training procedure described in~\cite{achterhold2021explore}.
For each training example, the context set size~$K$ is uniformly sampled in~$\{0,\ldots,50\}$. 
The target rollout length is $N=50$. 
As in EtC~\cite{achterhold2021explore}, we set $\lambda_{\mathit{KL}} = 5$. 

\subsection{Network architecture} \label{sec:network_architecture}

We set the dimensionality of the latent variable~$\vbeta$ to~$16$. 
The encoder MLPs~$e_{\beta}$,~$e_{\tau}$ and~$e_{u}$ contain a single hidden layer with 200 units and ReLU activations and an output layer with ReLU activation which maps to an embedding of dimensionality~$200$.
The encoder MLP~$e_{x}$ contains a single hidden layer with 200 units and ReLU activations and an output layer with tanh activation which maps to an embedding of dimensionality~$200$.
The MLPs of the state decoders~$d_{x,\mu}, d_{x,\sigma^2}$ consists of a single hidden layer with 200 units and ReLU activations. 
A linear output layer maps to the mean for~$d_{x,\mu}$.
For~$d_{x,\sigma^2}$, the linear output layer is followed by a softplus activation function.
The context encoder applies an MLP~$e_{\mathrm{trans}}$ to each state-action-state transition with one hidden layer with 200 units and a linear output layer with output dimensionality~$128$. 
Each layer is followed by a ReLU activation.
The outputs of the transition encoder are aggregated using a max-operation and decoded into mean and variance by MLPs $d_{\beta,\mu}, d_{\beta,\sigma^2}$.
Both MLPs contain one hidden layer with 200 units followed by a ReLU activation and a linear layer which maps to output dimensionality~$128$.
The output activation function of~$d_{\beta,\sigma^2}$ is a softplus activation and the weights are non-negative as described in~\cref{sec:background}.

\subsection{Model ablation} \label{sec:model_ablation}
As an ablation to our model, we only input the terrain features $\bm{\tau}$ at the current and previously visited states of the robot as terrain observations, but do not allow for terrain lookups  in a map at future states  during prediction. We will refer to this ablation as \emph{No($-$) terrain lookup} in the following.

\subsection{Prediction evaluation}
\label{sec:tradyn:eval_prediction}
In this section we evaluate the prediction performance of our proposed model. To this end, we generate 150 test trajectories of length 150, on the \emph{test} set of terrain layouts $\mathcal{A}_\mathrm{test}$. Robot parameters are uniformly sampled as during data collection for model training. The robot's initial position is sampled from ${[0.1, 0.9)}^2$, the orientation from $[0, 2\pi)$. The initial velocity is fixed to 0. Actions are sampled with a time-correlated (colored) noise scheme. In case the model is \emph{calibrated}, we additionally collect a small trajectory for each trajectory to be predicted, consisting of 10 transitions, starting from the same initial state $\bm{x_0}$, but with different random actions. Transitions from this trajectory form the context set $\sC$, which is used by the context encoder $\qctx(\bm{\beta} \mid \sC)$ to output a belief on the latent context variable~$\bm{\beta}$. In case the model is \emph{not calibrated}, the distribution is given by the context encoder for an empty context set, i.e. $\qctx(\bm{\beta} \mid \sC = \{\})$. We evaluate two model variants; first, our proposed model which utilizes the terrain map~${\tau}(\cpx, \cpy)$ for lookup during predictions, and second, a model for which the terrain observation is concatenated to the robot observation.
All results are reported on 5 independently trained models.
\Cref{fig:tradyn:unicycle_prediction} shows that our approach with terrain lookup and calibration clearly outperforms the other variants in position and velocity prediction. As the evolution of the robot's angle is independent of terrain friction, for angle prediction, only performing calibration is important.


\subsection{Planning evaluation}
\label{sec:tradyn:eval_planning}
Aside the prediction capabilities of our proposed method, we are interested whether it can be leveraged for efficient navigation planning. To evaluate the planning performance, we generate 150 navigation tasks, similar to the above prediction tasks, but with an additional randomly sampled target position $\bm{p}^* \in {[0.1,0.9)}^2$ for the robot. We perform receding horizon control as described in \cref{sec:tradyn:pathplanning}.

Again, we evaluate four variants of our model. We compare models with and without the ability to perform terrain lookups. Additionally, we evaluate the influence of calibration, by either collecting 10 additional calibration transitions for each planning task setup, or not collecting any calibration transition ($\sC = \{ \}$), giving four variants in total.

As we have trained five models with different seeds, over all models, we obtain 750 navigation results. We count a navigation task as \emph{failed} if the final Euclidean distance to the goal exceeds $\SI{5}{\centi\meter}$. 

We evaluate the efficiency of the navigation task solution by the sum of squared throttle controls over a fixed trajectory length of $N=50$ steps, which we denote as $E = \sum_{n=0}^{N-1} u_{\mathrm{throttle}, n}^2$.
We introduce super- and subscripts $E^v_{k,i}$ to refer to model variant $v$, planning task index $k$ and model seed $i$. Please see \cref{fig:tradyn:unicycle_planning_gain,fig:tradyn:unicycle_planning_examples} for results comparing the particular variants.  
Note that we filter out evaluation trajectories where any of the variants failed to reach the goal, since these can get stuck but have low control energy.
For pairwise comparison of control energies $E$ we leverage the Wilcoxon signed-rank test with a $p$-value of $0.05$.
We can conclude that, regardless of calibration, performing terrain lookups yields navigation solutions with significantly lower throttle control energy. The same holds for performing calibration, regardless of performing terrain lookups. Lowest control energy is obtained for both performing calibration and terrain lookup.

We refer to \cref{table:tradyn:planning_statistics} for statistics on the number of failed tasks.
As can be seen, our terrain- and robot-aware approach reaches the goal most often in all three threshold settings.
Planning with non-calibrated model variants fails much more often.
We show such failure cases in \cref{fig:tradyn:unicycle_planning_fail}.

\begin{table}[tb]
\centering
\caption{Failure counts for \SI{10}, \SI{5} and \SI{1}{\centi\meter} distance threshold to goal. Total number of tasks is 750.
Variants are with/without terrain lookup ($\pm$T) and with/without calibration ($\pm$C). Our full approach ($+$T, $+$C) yields best goal reaching performance.
}
\label{table:tradyn:planning_statistics}
\begin{tabular}{lccc}
    \toprule
    Variant & \multicolumn{3}{c}{Failed tasks} \\
    \cmidrule(lr){2-4}
    & \SI{10}{\centi\meter} & \SI{5}{\centi\meter} & \SI{1}{\centi\meter} \\
    \midrule
    {\scriptsize \color{mpl_propcycle_1} \newmoon} $-$T, $-$C & 5 & 7 & 112 \\
    {\scriptsize \color{mpl_propcycle_2} \newmoon} $-$T, $+$C & \textbf{0} & \textbf{0} & 30 \\
    {\scriptsize \color{mpl_propcycle_3} \newmoon} $+$T, $-$C & 5 & 12 & 72 \\
    {\scriptsize \color{mpl_propcycle_4} \newmoon} $+$T, $+$C & \textbf{0} & \textbf{0} & \textbf{2} \\
    \bottomrule
\end{tabular}
\end{table}

\subsection{Experiments with noise} \label{sec:noise_experiments}
In this section, we study the resilience of our method to stochastic dynamics and noisy data.
We consider two distinct forms of noise: \textit{action noise}, where the dynamics are stochastic, i.e., with $\sigma_a > 0$ but the model is given ground-truth observations, as well as \textit{observation noise}, where the dynamics are deterministic, but the model is presented with observations that have been corrupted by additive Gaussian noise, i.e. $\tilde{\bm{x}}(t) = \mathrm{clip}(\bm{x}(t) + \boldsymbol{\epsilon}_o)$ where $\boldsymbol{\epsilon}_o \sim \mathcal{N}(0, \mathbf{\Sigma}_o)$ and the $\mathrm{clip}$ function clips each component of the state vector to its boundaries as stated in \cref{sec:sim_robot_dynamics}.

The observation noise covariance $\boldsymbol{\Sigma}_o = \operatorname{diag}(\tilde{\sigma}_{o,i}^2)$ is diagonal.
To use intuitive values, we base each $\tilde{\sigma}_{o,i}$ on a single common $\sigma_o$ that is scaled to the maximum range of the respective observation component.
Concretely, $\tilde{\sigma}_{o,i} = \sigma_o (v_{\text{max},i} - v_{\text{min},i}) / 2$ where $v_{\text{max},i}, v_{\text{min},i}$ denote the boundaries of the $i$th observation component.
This means that e.g. for a component $i$ that has boundaries~$-1$ and $1$, $\tilde{\sigma}_{o,i} = \sigma_o$.
Furthermore, we distinguish two variants of observation noise:
One denoted \textit{physical}, which models imperfect state estimation and applies noise as described to all physical components of the observation, i.e. those introduced in \cref{sec:sim_robot_dynamics}, but leaves the observations of the terrain (its color value, see \cref{sec:tradyn:terrainlayouts}) at ground-truth values.
However, the terrain feature lookup will be performed with the noisy position and can therefore differ from the value at the robot's ground-truth position.
Our second observation noise variant, \textit{terrain}, uses a noisy terrain map but leaves physical features at ground truth.
To obtain the noisy map, Gaussian noise is added to each component of the terrain feature (i.e., its color channel) and at each position independently.
Note that since the terrain features are in the range $[0, 1]$, a standard deviation of $\sigma_o / 2$ is used for the noise.

\begin{figure}[tb!]
    \centering
    \includegraphics[scale=1]{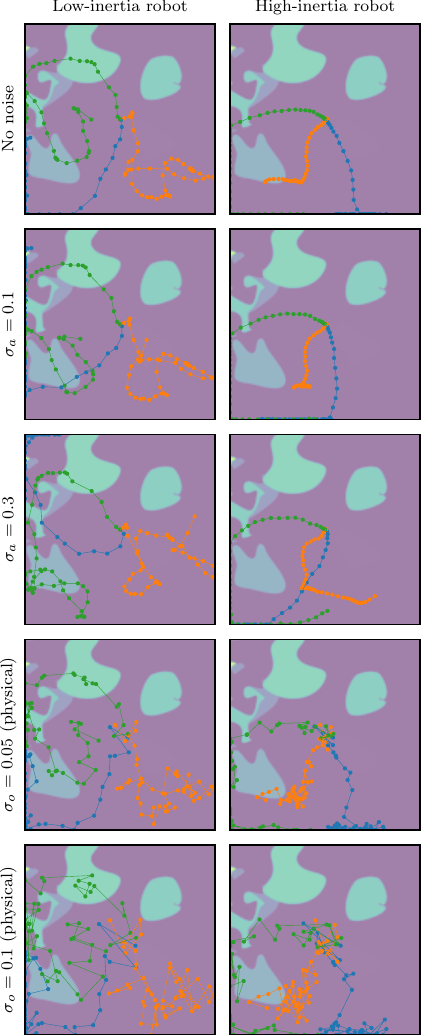}
    \caption{Exemplary rollouts (length 50) for two exemplary robot configurations and different types and amounts of noise.
        The first two rows have action, the last two rows observation noise.
        Same terrain, robot parameters and action sequences (for identical colors) as in \cref{fig:tradyn:rollouts} (terrain 1).
        Equally colored trajectories ({\scriptsize \color{mpl_propcycle_1} \newmoon}, {\scriptsize \color{mpl_propcycle_2} \newmoon}, {\scriptsize \color{mpl_propcycle_3} \newmoon}) correspond to identical sequences of applied actions.
        }
    \label{fig:tradyn:rollouts_noisy}
\end{figure}

\begin{figure}[tb!]
    \centering
    \includegraphics[scale=1]{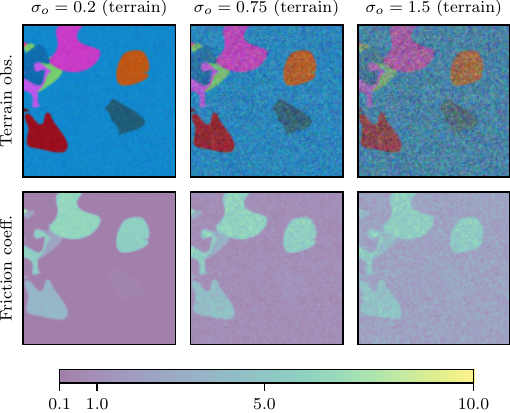}
    \caption{Exemplary noisy terrain maps. Terrain features and the respective calculated friction coefficient is shown for different levels of terrain observation noise.
        Same terrain as in \cref{fig:tradyn:terrains} (terrain 1).}
    \label{fig:tradyn:terrains_noisy}
\end{figure}

In \cref{fig:tradyn:rollouts_noisy}, we visualize some exemplary trajectories with different types and levels of noise applied.
Note that the same environment and identical action sequences are used for trajectories with the same color.
The noise-free case is identical to the first setting in \cref{fig:tradyn:rollouts}.
In \cref{fig:tradyn:terrains_noisy},  we apply observation noise to the terrain features of the environments from \cref{fig:tradyn:terrains} and show the resulting friction coefficient that corresponds to those observations.

For the following experiments, we trained all four model variants (with and without calibration, with and without terrain lookup) on datasets with different noise settings.
Training is conducted similar to the noise-free case as detailed in \cref{sec:model_training}, only that the trajectories are collected from environments with the respective noise parameters set.
We study each form of noise independently (action, physical observation, terrain observation noise), so each dataset only features one type and fixed level of noise.
In the following, we compare models trained on these different datasets.

\subsubsection{Prediction evaluation with noise} \label{sec:eval_pred_noisy}
We conduct experiments similar to \cref{sec:tradyn:eval_prediction} on the models trained with noise.
For each model variant (regarding calibration and terrain lookup), 5 independently trained models are evaluated on the same 150 test trajectories of length 150.
The test trajectories are obtained from environments with the same noise characteristics as those from which the training trajectories were collected.
This means that predictions from models trained with some action noise $\sigma_a > 0$ will be compared against stochastic ground-truth trajectories that result from the same level of action noise $\sigma_a$.
Models trained with obervation noise will be provided with a noisy initial observation from which they have to predict future states, obtained by applying similar levels of observation noise $\sigma_o$ as seen on the training trajectories.
For the error calculation however, we assume that we have access to the ground-truth robot state.
This models the case where a model has to deal with stochastic dynamics or noisy sensor readings, but we can obtain an accurate estimate of the robot's position and velocity for evaluation, e.g. through external measurements.
Calibration trajectories are likewise obtained from environments with identical noise characteristics.

\begin{figure}[tb!]
    \centering
    \includegraphics[width=\linewidth]{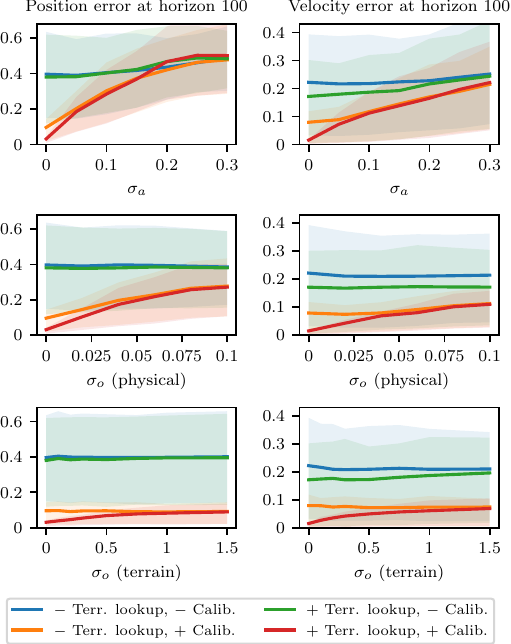}\\
    \caption{Prediction error for the proposed model and its ablations, plotted over different levels of noise.
        The same metrics as in \cref{fig:tradyn:unicycle_prediction} are used (Euclidean distance of the positional, absolute difference of the velocity).
        The prediction horizon is fixed at 100.
        Depicted are the mean and 20\%, 80\% percentiles over 150 evaluation rollouts for 5 independently trained models per model variant.
        }
    \label{fig:tradyn:unicycle_prediction_over_noise}
\end{figure}

Instead of plotting the prediction error over the prediction horizon as in \cref{fig:tradyn:unicycle_prediction}, we consider a fixed horizon of 100 and compare the models' prediction performance under different noise conditions in \cref{fig:tradyn:unicycle_prediction_over_noise}.
Regarding velocity estimation, we find that the results from \cref{sec:tradyn:eval_prediction} still hold up to high noise levels:
Each model with calibration outperforms its counterpart without calibration and each model with terrain lookup outperforms its counterpart without it.
For high noise levels however, the performance of the proposed model ($+$T, $+$C) approaches that of the ablation without terrain lookup ($-$T, +C). 
Regarding the position estimation error, results are similar, except for both variants without calibration performing equally well and the fact that from around $\sigma_a = 0.15$ and higher, the proposed variant ($+$T, $+$C) performs worse than the ablation without terrain lookup ($-$T, $+$C).
For terrain observation noise specifically, we observe an expected effect: With increasing levels of noise, the performance of variants with terrain lookup converge to that of their counterparts without lookup.
Otherwise, terrain noise does not significantly affect the prediction performance.

We conclude that for forward prediction, our results found in earlier sections also hold on stochastic environments with action noise and on deterministic environments with noise added to the observation.
With increasing amounts of noise however, the benefit of terrain lookup becomes less prevalent.
The benefit of calibration stays significant even up to high noise levels.

\subsubsection{Planning evaluation with noise} \label{sec:eval_plan_noisy}

\begin{figure}[tb!]
    \centering
    \includegraphics[width=\linewidth]{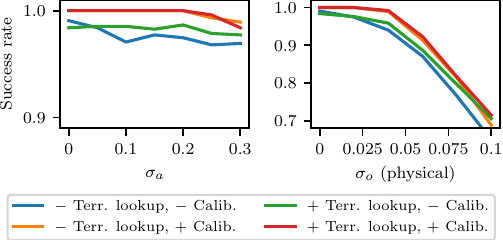}\\
    \caption{Task success rate for all model variants plotted over noise level.
        This metric corresponds to the failed tasks from \cref{table:tradyn:planning_statistics} at \SI{5}{\centi\meter} (one minus the quotient).
        }
    \label{fig:tradyn:success_rate_over_noise}
\end{figure}

\begin{figure*}[tb!]
    \centering
    \begin{subfigure}{.5\linewidth}
        \centering
        \includegraphics[width=.85\linewidth]{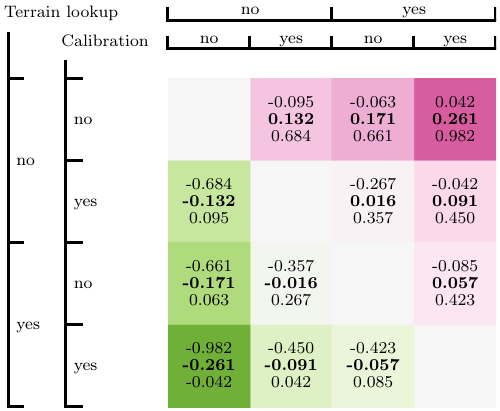}
        \caption{action noise $\sigma_a = 0.05$, 725 samples}
        \label{fig:gain_comp_act_low}
    \end{subfigure}%
    \hfill%
    \begin{subfigure}{.5\linewidth}
        \centering
        \includegraphics[width=.85\linewidth]{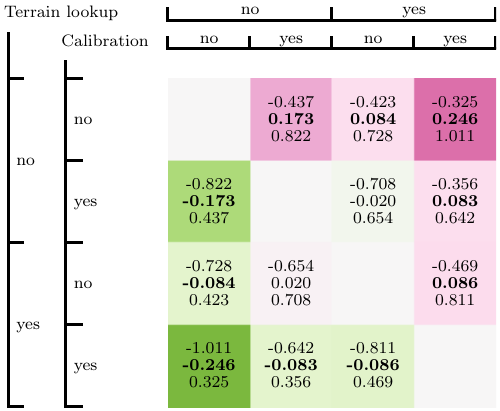}
        \caption{action noise $\sigma_a = 0.2$, 700 samples}
        \label{fig:gain_comp_act_high}
    \end{subfigure}\\[4ex]
    \begin{subfigure}{.5\linewidth}
        \centering
        \includegraphics[width=.85\linewidth]{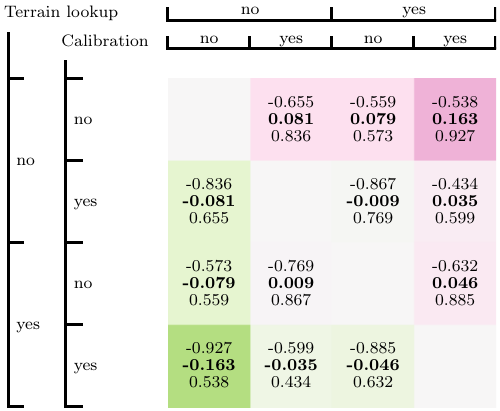}
        \caption{observation noise $\sigma_o = 0.02$ (physical), 705 samples}
        \label{fig:gain_comp_obs-phys_low}
    \end{subfigure}%
    \hfill%
    \begin{subfigure}{.5\linewidth}
        \centering
        \includegraphics[width=.85\linewidth]{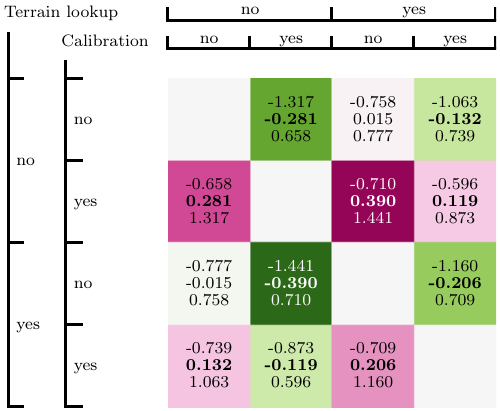}
        \caption{observation noise $\sigma_o = 0.06$ (physical), 390 samples}
        \label{fig:gain_comp_obs-phys_high}
    \end{subfigure}\\[4ex]
    \begin{subfigure}{.5\linewidth}
        \centering
        \includegraphics[width=.85\linewidth]{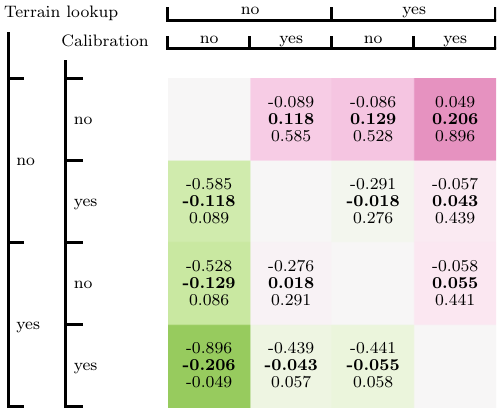}
        \caption{observation noise $\sigma_o = 0.5$ (terrain), 680 samples}
        \label{fig:gain_comp_obs-terr_low}
    \end{subfigure}%
    \hfill%
    \begin{subfigure}{.5\linewidth}
        \centering
        \includegraphics[width=.85\linewidth]{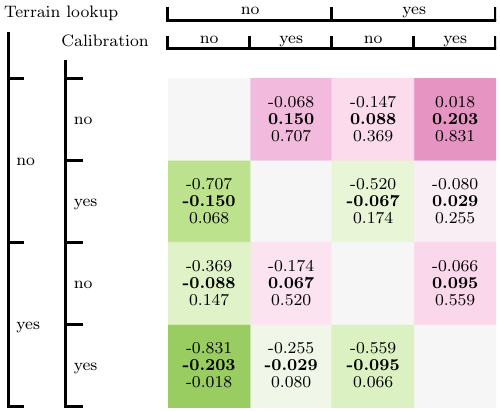}
        \caption{observation noise $\sigma_o = 1.0$ (terrain), 670 samples}
        \label{fig:gain_comp_obs-terr_high}
    \end{subfigure}
    \caption{Control cost comparison matrices between model variants with and without terrain lookup and calibration for different noise settings.
        The same metric as for \cref{fig:tradyn:unicycle_planning_gain} is used, i.e. throttle control energy differences between the variants on navigation task $k \in \{1, \ldots, 150\}$.
         We show statistics (20\% percentile, median, 80\% percentile) on the set of pairwise comparisons of control energies $\{ 
    E^{\mathrm{row}}_{k,i_1} - E^{\mathrm{col}}_{k,i_2} \mid \forall k \in \{1,\ldots,K\}, i_1 \in \{1,\ldots,5\}, i_2 \in \{1,\ldots,5\}
    \}$.
        Significant ($p<0.05$) results are printed \textbf{bold} (see \cref{sec:tradyn:eval_planning}).
        Different sample sizes are a result of filtering the evaluation trajectories to instances where each variant reached the goal.
        }
    \label{fig:tradyn:unicycle_planning_gain_noisy}
\end{figure*}

Similarly to \cref{sec:tradyn:eval_planning}, we evaluate the performance of our trained models in a planning scenario, but now on noisy environments.
Similarly to the previous \cref{sec:eval_pred_noisy}, we evaluate the models on environments that exhibit the same noise characteristics as seen during training.
This means that models trained on stochastic trajectories with some $\sigma_a > 0$ now have to perform planning in an environment with the same level of action noise $\sigma_a$.
Likewise, models trained on noisy observations originating from some $\sigma_o$ are now provided with similarly noisy observations from which they have to plan.

The success rates of the variants reaching the target (less than 5cm distance to goal) are visualized in \cref{fig:tradyn:success_rate_over_noise}.
For the case of action noise, both variants with calibration maintain a very high success rate almost up to the maximum noise level considered.
At $\sigma_a = 0.3$, the ($-$T, $+$C) variant failed to reach only 8 / 750 targets and the ($+$T, $+$C) variant 12 / 750.
Regarding physical observation noise, the ($+$T, $+$C) variant maintains the highest success rate until $\sigma_o = 0.04$.
With increasing noise levels, all variants experience a comparable drop in performance.
We found that terrain observation noise did not significantly impact the performance of calibrated variants, which is why we do not show a plot over terrain noise.
We conclude that all model variants with calibration are very resilient to action noise when used for planning.
For physical observation noise, there is a noise level up to which results are similar to the noise-free case, but after which planning performance is seriously deteriorated for all variants with increasing noise level.

In \cref{fig:tradyn:unicycle_planning_gain_noisy}, we show control cost comparison matrices similar to \cref{fig:tradyn:unicycle_planning_gain} for exemplary noise levels.
For all three types of noise, we show one case with lower and one with higher noise value.
Note that we only consider trajectories where all ablation variants (regarding calibration and terrain lookup) reached the goal.
For action noise (\cref{fig:gain_comp_act_low} and \cref{fig:gain_comp_act_high}), we can see a trend that increases with noise:
When comparing with variants that use either calibration or terrain lookup, the noise-free case (\cref{fig:tradyn:unicycle_planning_gain}) prefers terrain lookup over calibration, whereas for high action noise, they perform almost equally well.
A similar effect can be observed for the small physical observation noise value (\cref{fig:gain_comp_obs-phys_low}).
With stronger physical observation noise however (\cref{fig:gain_comp_obs-phys_high}), variants with calibration even perform worse in terms of control energy than those without.
For the case of terrain observation noise (\cref{fig:gain_comp_obs-terr_low} and \cref{fig:gain_comp_obs-terr_high}), a similar effect as for action noise can be observed:
When comparing terrain lookup with calibration, with increasing noise level, the preference shifts towards calibration.
Note that the control cost comparison is only relevant when the goal is reached, and even if a variant compares favorably in terms of control energy, it may have a lower success rate as depicted in \cref{fig:tradyn:success_rate_over_noise}.
We conclude that in terms of control energy, for all cases except high physical observation noise, our proposed method with calibration and terrain lookup outperforms its ablations with statistical significance.
Regarding the ablations, calibration generally becomes more important than terrain lookup with increasing noise levels.
An exception is at a high physical observation noise level, where the variant with only terrain lookup enabled compares best to all other variants.

\subsection{Discussion and Limitations}

The proposed approach assumes that the robot state is fully observable and that a map of terrain features is known a priori.
In future work, the former assumption could be alleviated by using a state estimator or learning a filter concurrently with the dynamics model, similar to the approach in~\cite{hafner2019learning}.
For the experiments with noise, we assumed independent additive Gaussian noise, both for noisy execution of actions and state observations.
Thereby, we did not assess different systematic sources of error such as outliers.
Especially for the map, the noise contained in the stored terrain features might be more complex than independent Gaussian noise due to the measuring and fusion process.
An interesting avenue for future research is also to learn the terrain feature map from data while navigating through the environment.

\section{Conclusions}
In this paper, we propose a forward dynamics model which can adapt to variations in unobserved variables that govern the system's dynamics such as robot-specific properties as well as to {spatial} variations. We train our model on a simulated unicycle-like robot, which has varying mass and actuator gains. In addition, the robot's dynamics are influenced by instance-wise and spatially varying friction coefficients of the terrain, which are only indirectly observable through terrain observations. 
In 2D simulation experiments, we demonstrate that our model can successfully cope with such variations through calibration and terrain lookup. It exhibits smaller prediction errors compared to model variants without calibration and terrain lookup and yields solutions to navigation tasks which require lower throttle control energy. 
We also assess the prediction and planning performance of our approach under various noise sources and evaluate its resilience to various levels of noise.
Our results demonstrate that for moderate noise levels in dynamics, physical state observations and terrain features, calibration and terrain look-up are still beneficial in our model.
In future work, we plan to extend our novel learning-based approach for real-world robot navigation problems with partial observability.

\section*{Author contributions}
\textbf{Jan Achterhold}: Conceptualization, Methodology, Software, Validation, Formal analysis, Investigation, Writing - Original Draft, Writing - Review \& Editing, Visualization.
\textbf{Suresh Guttikonda}: Conceptualization, Methodology, Software, Formal analysis, Writing - Original Draft, Writing - Review \& Editing.
\textbf{Jens Kreber}: Conceptualization, Software, Validation, Formal analysis, Investigation, Writing - Original Draft, Writing - Review \& Editing, Visualization.
\textbf{Haolong Li}: Conceptualization, Methodology, Writing - Review \& Editing.
\textbf{Joerg Stueckler}: Conceptualization, Methodology, Formal analysis, Writing - Original Draft, Writing - Review \& Editing,  Supervision, Project administration, Funding acquisition.

\section*{Acknowledgements}

This work has been supported by Max Planck Society and Cyber Valley. The authors thank the International
Max Planck Research School for Intelligent Systems (IMPRS-IS) for supporting Jan Achterhold and Haolong Li.

\balance



\bibliographystyle{elsarticle-num} 
\bibliography{bibliography}

\end{document}